\documentclass{iopjournal}

\usepackage{cite}
\usepackage{amsmath,amssymb,amsfonts}
\usepackage{algorithmic}
\usepackage{textcomp}
\usepackage{booktabs}
\usepackage{multirow,multicol}
\usepackage{graphicx}
\usepackage{subcaption}

\usepackage{array}

\newcolumntype{H}{>{\setbox0=\hbox\bgroup}c<{\egroup}@{}}

\newcommand{\spacedhline}{\hline \rule{0pt}{2ex}}

\begin{document}

\articletype{Topical Review} %	 e.g. Paper, Letter, Topical Review...

\title{Cross-Subject Generalization for EEG Decoding: A Survey of Deep Learning Methods}
% Cross-Subject EEG Generalization with Deep Learning: A Survey

\author{Taida Li$^1$\orcid{0009-0009-0037-3345}, Yujun Yan$^2$\orcid{0000-0003-3776-4293}, Fei Dou$^{3}$\orcid{0000-0003-4246-8616}, Wenzhan Song$^{4}$\orcid{0000-0001-8174-1772} and Xiang Zhang$^{1, *}$\orcid{0000-0001-5097-2113}}

\affil{$^1$Department of Computer Science, University of North Carolina at Charlotte, Charlotte, United States}

\affil{$^2$Department of Computer Science, Dartmouth College, Hanover, United States}

\affil{$^3$School of Computing, University of Georgia, Athens, United States}

\affil{$^4$School of Electrical and Computer Engineering, University of Georgia, Athens, United States}

\affil{$^*$Author to whom any correspondence should be addressed.}

\email{xiang.zhang@charlotte.edu}


\begin{abstract}
Deep learning for cross-subject EEG decoding is hindered by high inter-subject variability, which introduces a severe domain shift between training and unseen test subjects. This survey presents a comprehensive review of deep learning methodologies specifically engineered to address this cross-subject generalization challenge. To ground this analysis, we formalize the cross-subject setting as a multi-source domain problem and delineate the rigorous, subject-independent evaluation protocols required for valid assessment. Central to this survey is a systematic taxonomy of the current literature into discrete methodological families, including feature alignment, adversarial learning, feature disentanglement, and contrastive learning. We conclude by examining three critical elements for advancing robust, real-world decoding: the theoretical limitations of current methodologies, the structural value of subject identity, and the emergence of EEG foundation models.
\end{abstract}

\section{Introduction}
\label{sec:intro}
\begin{figure}[t]
    \centering
    \includegraphics[width=0.9\linewidth]{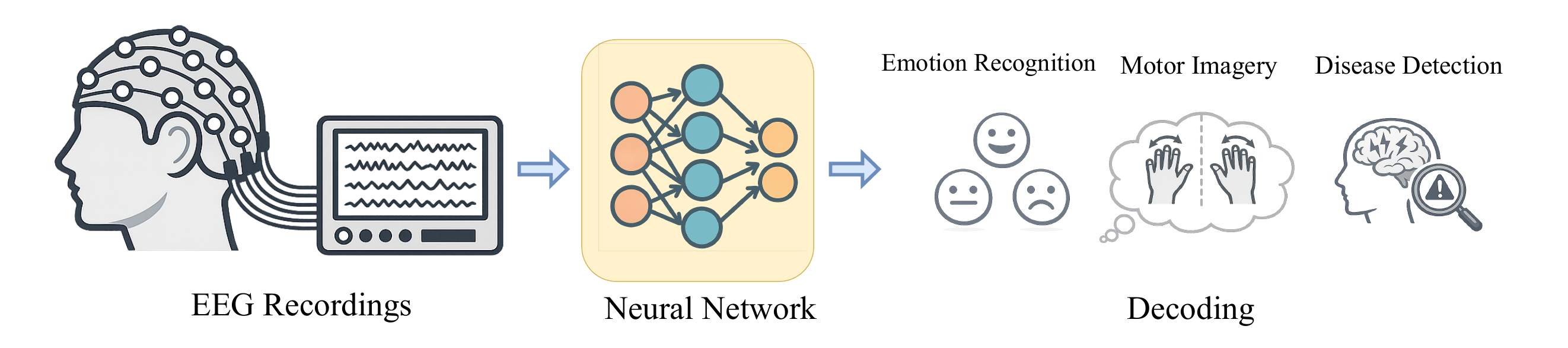}
    \vspace{-2mm}
    \caption{Deep Learning Pipeline for EEG Decoding.}
    \vspace{-3mm}
    \label{fig:eeg_schemes}
\end{figure}

The application of deep learning to electroencephalography (EEG) signal decoding has marked a significant paradigm shift in computational neuroscience and brain-computer interfaces (BCIs)~\cite{Craik2019}. Due to their capacity for automatic feature extraction from high-dimensional time-series data, deep neural networks have emerged as a promising paradigm complementing traditional machine learning pipelines that rely on hand-crafted features. This has led to notable advancements in a variety of applications, including clinical diagnostics for conditions like epilepsy~\cite{ACHARYA2018, khan2021epileptic}, analysis of cognitive and affective states such as emotion recognition~\cite{zheng2015emotion, wagh2023review}, and the decoding of motor imagery~\cite{lawhern2018eegnet}. A schematic of this end-to-end process is illustrated in Fig.~\ref{fig:eeg_schemes}, demonstrating how raw EEG recordings are mapped through neural network feature extractors to generate these diverse downstream predictions.

Despite these successes, a fundamental obstacle hinders the translation of these models from laboratory settings to practical, real-world applications: the profound inter-subject variability of EEG signals~\cite{haegens2014variability, saha2019variability}. Individual variations in physiology, anatomy, and cognition manifest as distinct neural signatures, leading to significant inter-subject variability~\cite{saha2019variability}. This high variability has two critical consequences for deep learning models. First, it creates a domain shift, where the data distribution of a new, unseen subject is significantly different from that of the subjects in the training set. Second, a high-capacity neural network is prone to overfitting to the salient, subject-specific features rather than learning the more subtle, task-relevant neural patterns. The combination of this domain shift and the model's tendency to exploit subject-specific confounds leads to a catastrophic drop in performance when generalizing to a new user, which is the central cross-subject challenge. This situation presents a unique, two-sided problem. The challenge is rooted in the powerful, subject-specific biomarkers inherent in the EEG signals, to which models are prone to overfit. The opportunity, however, arises because this inter-subject variability is not random noise, but rather a structured phenomenon tied to the individual. Physiological datasets typically possess rich metadata—specifically the knowledge of which subject generated which signals. This metadata allows researchers to design cross-subject methodologies that model, align, or disentangle the very variability that makes generalization difficult.

The unique character and challenge of cross-subject generalization have led researchers to propose methodologies designed to explicitly address it, which forms the scope of this survey. These approaches directly confront inter-subject variability by strategically utilizing the structural information available in the dataset. Feature Alignment frameworks aim to minimize the distribution shift between source subjects and a specific target subject, often through statistical moment matching or geometric alignment. Adversarial Learning paradigms introduce a ``minimax'' game, training feature extractors to fool a subject discriminator, thereby enforcing the learning of subject-invariant representations. Feature Disentanglement approaches go a step further, mathematically decomposing the neural signal into distinct task-relevant and subject-specific components. Contrastive Learning methods leverage metadata to structure the embedding space, defining positive and negative pairs to either cluster data by task across subjects or explicitly separate subject identities. Meta-Learning reformulates the training process itself, using episodic training to simulate domain shift and optimize models for rapid adaptability. Invariant and Causal Representation Learning methods seek to discover stable causal mechanisms that remain constant across diverse patient environments, disregarding spurious subject-specific correlations. We provide a detailed categorization and in-depth analysis of these diverse methodological families in Section \ref{sec:method}.

In contrast, we consider works that adopt a subject-independent evaluation but do not contain a specific mechanism to leverage subject information as being outside the scope of this survey. Such approaches, which may include the design of more powerful pre-training methods~\cite{zhang2022self} or more capable generic encoders~\cite{wang2024medformer,fan2025medgnn}, are contributions to the general representation learning for medical time series rather than targeted solutions to the cross-subject generalization problem. This survey will proceed to categorize and analyze the families of methods that explicitly harness subject-level information to learn robust and generalizable representations from EEG signals.

While previous reviews have recognized the challenge of cross-subject generalization, they typically restrict their focus to isolated application domains, such as emotion recognition \cite{apicella2024toward} or seizure detection \cite{shafiezadeh2024systematic}. Furthermore, these existing reviews tend to provide a generalized overview encompassing both traditional machine learning and deep learning frameworks. The current survey distinguishes itself through a dual approach: we expand the application scope to encompass a diverse range of tasks, including emotion recognition, motor imagery, and broader disease detection, while simultaneously narrowing our technical lens exclusively to deep learning paradigms. By focusing exclusively on deep learning, this survey offers a more technical analysis, categorizing works by their core methodologies rather than general learning settings.

\section{Background}
\label{sec:background}
To systematically survey the landscape of cross-subject methodologies, it is essential to first establish a clear definitional framework. This section will formalize the cross-subject generalization problem using machine learning terminology, clarify the terminology used throughout the field, define the principles of rigorous evaluation, and introduce the related research areas that provide the methodological toolkit for the techniques discussed later in this paper.

\subsection{Terminology}

The academic literature addresses the challenge of generalizing models to new individuals using a variety of terms. Phrases such as ``cross-subject,'' ``cross-patient,'' ``subject-independent,'' and ``subject-invariant'' are often used interchangeably to describe the same fundamental goal: creating a model that performs robustly on an individual whose data was not seen during training. For the sake of consistency and generality, this survey will adopt the term \textbf{cross-subject} to refer to this general problem and the methodologies designed to solve it.

\subsection{Inter-Subject Variability in EEG}
\label{sub:variability}

The fundamental challenge in cross-subject EEG decoding, which we formally define as a domain shift, stems from the ``inter-subject variability." This is the observation that EEG signals are not uniform across people, even when performing the same mental task. From a physiological perspective, this variability is expected; individual differences in brain anatomy, skull thickness, baseline neural rhythms, and even the cognitive strategies employed to perform a task all contribute to unique neural signatures.

This hypothesis is strongly supported by empirical evidence from deep-learning-based EEG decoding. The presence of these unique signatures is so pronounced that they act as biometric identifiers. Several studies demonstrate that when a standard deep learning model is trained on a multi-subject dataset, it can learn to identify the \textit{subject} with high accuracy. For instance, Özdenizci et al.~\cite{ozdenizci2020learning} observed that a standard CNN could achieve subject identification accuracy as high as 62.6\% on a 40-subject task (where chance is 2.5\%). Zhang et al.~\cite{zhang2024cross} similarly showed that a normally-trained model's "Identity Accuracy" progressively increases during training, proving that the model actively learns these subject-specific features. This is also visually confirmed by Shen et al.~\cite{shen2023contrastive}, who showed that in an unaligned embedding space, EEG features cluster by \textit{subject} rather than by emotional state. Empirical evidence confirms that the data distribution from one subject is significantly different from another.

\begin{figure}[!htbp]
    \centering
    % Second Subfigure
    \begin{subfigure}[b]{0.48\linewidth}
        \centering
        \includegraphics[width=\linewidth]{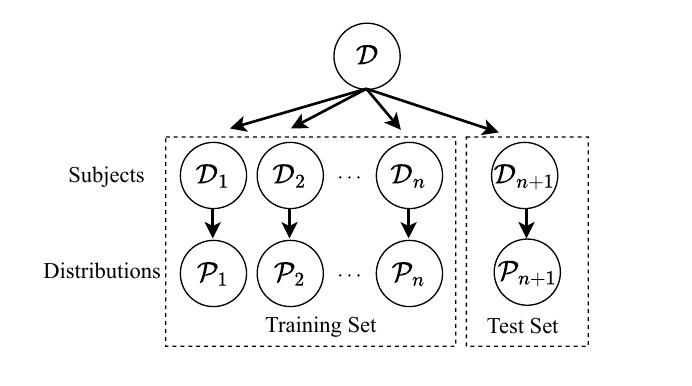}
        \caption{Cross-subject Setup} 
        \label{fig:cross_subject}
    \end{subfigure}
    \hspace{-3mm} % Adds flexible space between the images
    % First Subfigure
    \begin{subfigure}[b]{0.48\linewidth}
        \centering
        \includegraphics[width=\linewidth]{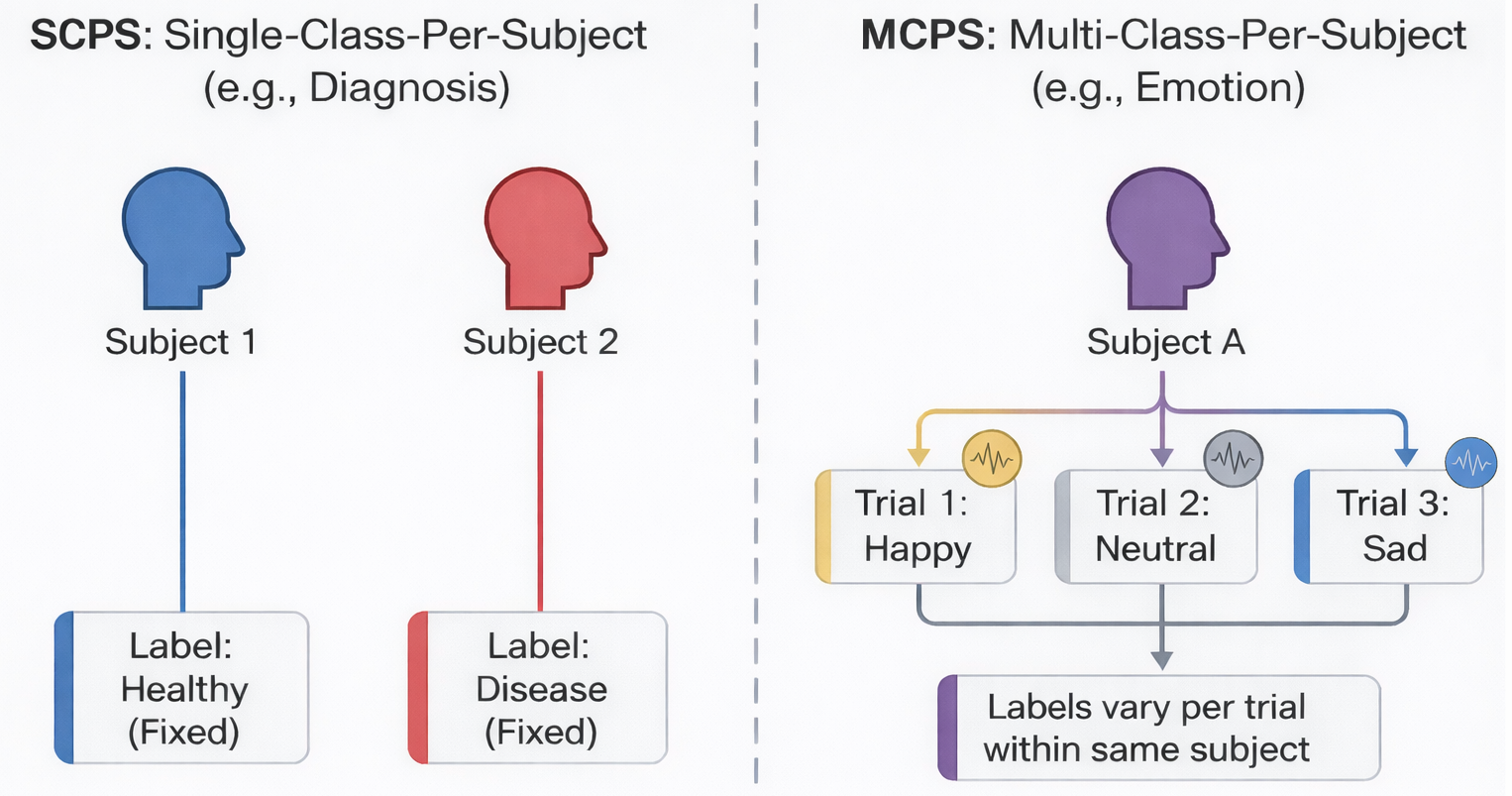}
        \caption{SCPS vs. MCPS} 
        \label{fig:scps_comparison}
    \end{subfigure}
    \caption{Problem formulation and task taxonomy for cross-subject EEG decoding. Panel (a) formalizes the cross-subject learning setting as a multi-source domain problem, illustrating the inherent distributional shift between the heterogeneous pool of training subjects and the disjoint, unseen target subject. Panel (b) categorizes classification tasks into Single-Class-Per-Subject (SCPS) and Multi-Class-Per-Subject (MCPS), demonstrating the structural differences in how clinical or cognitive labels map to individual subject identities.}
    \vspace{-3mm}
    \label{fig:eeg_analysis_main}
\end{figure}

\subsection{Problem Statement}
\label{sub:problem_statement}
The cross-subject generalization challenge is fundamentally a problem of real-world domain shift. Let the input space of EEG signals be denoted by $\mathcal{X}$ and the label space of cognitive or clinical states be $\mathcal{Y}$. We consider the data from each individual subject, $i$, as a distinct \textbf{domain}, characterized by a unique joint probability distribution $P_i(\mathbf{x}, y)$ over $\mathcal{X} \times \mathcal{Y}$. Due to the high inter-subject variability in EEG, the data distributions for any two different subjects are not identical, meaning $P_i \neq P_j$.

The core objective is to learn a predictive function $f: \mathcal{X} \to \mathcal{Y}$ using data from a finite set of observed source subjects, $S_{\text{source}}$, that minimizes the expected error on a novel, unobserved subject $k \notin S_{\text{source}}$.

Crucially, the challenge of domain shift is not limited to the transition to the novel subject; it is intrinsic to the observed data itself. Because $S_{\text{source}}$ consists of multiple individuals, it cannot be modeled as a single coherent distribution. Instead, it is a collection of heterogeneous domains where $P_i \neq P_j$ for any distinct subjects $i, j \in S_{\text{source}}$. This internal variability formally casts the cross-subject task as a multi-source domain problem, rather than a standard single-source transfer learning problem. This distinction provides the theoretical basis for applying advanced methodological frameworks, such as Multi-Source Domain Adaptation and Domain Generalization.

\subsection{Evaluation Protocols}
\label{sub:evaluation_protocols}
Validating a model's ability to solve the cross-subject generalization problem requires an experimental protocol that accurately simulates the real-world application to an unseen subject. In the literature, experimental setups are broadly categorized into two types: \textbf{Subject-Dependent} and \textbf{Subject-Independent} \cite{fazli2009subject, wang2024medformer}.

\subsubsection{Subject-Dependent Evaluation}
In a subject-dependent (or segment-based) evaluation, EEG signals from all subjects are pooled together, segmented into shorter windows, and randomly split into training and testing sets. Consequently, segments from the same subject appear in both the training and test splits. While common in general machine learning, this protocol is fundamentally flawed for clinical EEG applications because it fails to simulate a novel user and introduces severe data leakage. Because EEG signals contain strong subject-specific biometric signatures, a model can achieve high accuracy by memorizing the identity of the subject rather than learning the pathological patterns of the disease \cite{wang2024medtsevaluation, brookshire2024data}. Recent studies have demonstrated that models evaluated this way often suffer from massive performance inflation, dropping from near-perfect accuracy (e.g., $>95\%$) to much lower accuracy (e.g., $60\%$) when tested on unseen subjects \cite{brookshire2024data}.

\subsubsection{Subject-Independent Evaluation}
Subject-independent (or subject-based) evaluation is the experimental design that correctly mirrors the cross-subject generalization problem. It simulates the deployment to a novel domain by ensuring that the training and testing sets are strictly disjoint at the subject level. Let $S$ be the set of all subjects in a given dataset. This protocol partitions the dataset such that the set of training subjects, $S_{\text{train}}$, and testing subjects, $S_{\text{test}}$, satisfy $S_{\text{train}} \cap S_{\text{test}} = \emptyset$.

Under this protocol, training and test sets must remain strictly disjoint at the subject level. By withholding the target distribution during training, this evaluation exposes if a model relies on identity-based shortcuts that does not generalize to new subjects. Common implementations of the subject-independent evaluation protocol include Leave-One-Subject-Out (LOSO) cross-validation, group K-fold cross validation and fixed hold-out sets. Since it accurately simulates deployment to a novel domain, a subject-independent evaluation is not merely a rigorous testing mechanism, but a necessary structural reflection of the cross-subject problem itself. For clinical translation, subject-independent evaluation is the only valid metric of a model's utility to unseen subjects~\cite{wang2024medtsevaluation, brookshire2024data}.

\subsection{Label Structure of Classification Tasks}
\label{sub:scps_vs_mcps}

The nature of inter-subject variability manifests differently depending on the label structure of the classification task. As shown in Fig.~\ref{fig:scps_comparison}, we categorize EEG tasks into two distinct types based on the mapping between subjects and labels: \textbf{Single-Class-Per-Subject (SCPS)} and \textbf{Multi-Class-Per-Subject (MCPS)}.

\subsubsection{Single-Class-Per-Subject (SCPS)}
In SCPS tasks, each subject is assigned a single, fixed label that does not change over time. This is typical for disease diagnosis tasks such as Alzheimer's Disease (AD), Schizophrenia, or Major Depressive Disorder detection, where a subject is categorized as either ``Healthy" or ``Diseased" \cite{wang2024medtsevaluation}.

\textbf{Challenge: The Identity Shortcut.} The primary challenge in SCPS tasks is the perfect correlation between \textit{Subject Identity} and the \textit{Task Label} within the training set. Since the label is constant for a given subject, identifying the subject and then memorizing that subject's label is functionally equivalent to determining the label. Deep learning models, behaving as ``lazy'' learners, often exploit this by learning a trivial mapping from the subject's strong biometric signature (identity features) to the label. This strategy relies purely on memorization; consequently, when the model encounters an unseen subject in the test set, it detects a novel identity signature for which it has established no correlation with any task label, causing this identity-based decision rule to fail. However, it is important to qualify that models rarely rely \textit{solely} on this shortcut; empirical results on unseen subjects often exceed random chance, indicating that the model does capture and utilize some subject-invariant task features.

% We propose that the performance drop observed during testing may arise from the classifier's joint use of both feature types: during training, the model likely assigns non-negligible weights to the salient identity features alongside task features. When applied to unseen subjects, the encoder continues to extract identity features, but because the specific correlation between the new subjects' identities and the task label has not been learned, these identity features effectively function as noise. This noise obscures the decision boundary derived from the task features, leading to poor generalization on new subjects. 

% Furthermore, this issue is compounded by a subtle conceptual overlap in SCPS: because a subject's pathological state is constant, the ``task feature'' (e.g., having Alzheimer's) effectively becomes a subset of the ``identity feature'' (distinguishing Subject A from Subject B), making their disentanglement particularly challenging.

\subsubsection{Multi-Class-Per-Subject (MCPS)}
In MCPS tasks, a single subject experiences multiple class states over time. Examples include Seizure Prediction (where a subject transitions between inter-ictal and pre-ictal states), Emotion Recognition (transitioning between happy, sad, neutral), and Motor Imagery.

\textbf{Challenge: Distribution Shift.} In MCPS, the identity shortcut is not available because knowing "who" the subject is does not automatically reveal "what" state they are in. Instead, these tasks still suffer heavily from the domain shift caused by inter-subject variability. The neural manifestation of a "seizure" or "happiness" varies significantly in topology, frequency, and amplitude from person to person. Consequently, a model trained on Source Subject A may learn a decision boundary that is misaligned for Target Subject B, leading to poor generalization even if the model is trying to learn task-relevant features \cite{brookshire2024data}. While SCPS tasks struggle with \textit{spurious correlations} (identity predicting label), MCPS tasks struggle with \textit{conditional distribution shifts} (the expression of the label varying by identity).

To ground these theoretical distinctions in empirical practice, Table~\ref{tab:public_eeg_datasets} compiles the standard public EEG datasets utilized by the methodologies reviewed in this survey. By explicitly mapping each dataset to its corresponding SCPS or MCPS structure, the table illustrates how different clinical and cognitive applications inherently dictate the specific type of cross-subject challenge researchers must address.

\begin{table*}[ht]
\centering
\resizebox{\textwidth}{!}{%
\begin{tabular}{l|l|c|c|c|c|c|l}
\spacedhline
\textbf{Task} & \textbf{Dataset} & \textbf{\# Subj} & \textbf{Hz} & \textbf{Ch} & \textbf{\# Classes} & \textbf{Label Structure} & \textbf{Methods}\\ \spacedhline
\multirow{8}{*}{\textbf{Motor Imagery}} 
 & PhysioNet MI~\cite{schalk2004bci2000} & 109 & 160 & 64 & 4 & MCPS & \cite{xu2020cross}, \\
 & BCI Competition IV 2b~\cite{leeb2008bci} & 9 & 250 & 3 & 2 & MCPS & \cite{xu2020cross},~\cite{song2023global}\\
 & GigaScience MI~\cite{cho2017eeg} & 52 & 512 & 64 & 2 & MCPS & \cite{xu2020cross},~\cite{ozdenizci2020learning}\\
 & OpenBMI~\cite{lee2019eeg} & 54 & 1000 & 62 & 2 & MCPS & \cite{zheng2025cross},~\cite{kobler2022spd},~\cite{kwak2023subject},~\cite{ng2024subject}\\
 & BCI Competition IV 2a~\cite{brunner2008bci, tangermann2012review} & 9 & 250 & 22 & 4 & MCPS & \cite{hang2019cross},~\cite{zheng2025cross},~\cite{kobler2022spd},~\cite{xu2020cross},~\cite{jiang2024deep},~\cite{song2023global},~\cite{feng2022classification},~\cite{kwak2023subject},~\cite{ng2024subject},~\cite{zhang2023subject}\\
 & Stieger2021~\cite{stieger2021continuous} & 62 & 1000 & 64 & 2 or 4 & MCPS & \cite{kobler2022spd}\\
 & Yi2014~\cite{yi2014}& 10 & 200 & 60 & 4 & MCPS & \cite{xu2020cross}\\
 & BCI Competition III IVa~\cite{dornhege2004boosting} & 5 & 1000 & 118 & 2 & MCPS & \cite{hang2019cross}\\
\spacedhline
\multirow{4}{*}{\textbf{Emotion Recognition}} 
 & OVPD-II~\cite{xue2022ovpd} & 13 & 250 & 28 & 3 & MCPS & \cite{zhang2023cross} \\
 & SEED-IV~\cite{8283814} & 15 & 200 & 62 & 4 & MCPS & \cite{yu2025fmlan},~\cite{she2023multisource} \\
 & SEED~\cite{zheng2015investigating, duan2013differential},~\cite{yu2025fmlan} & 15 & 200 & 62 & 3 & MCPS & \cite{she2023cross},~\cite{jiang2024deep},~\cite{bao2021two},~\cite{hwang2020subject},~\cite{alameer2024cross},~\cite{shen2023contrastive},~\cite{she2023multisource},~\cite{ahmed2023novel},~\cite{li2022cross},~\cite{liu2025mixeeg} \\
 & DEAP~\cite{koelstra2011deap} & 32 & 512 & 32 & 9 or 5 & MCPS & \cite{she2023cross},~\cite{alameer2024cross},~\cite{she2023multisource},~\cite{ahmed2023novel},~\cite{li2022cross},~\cite{bhosale2022calibration} \\
\spacedhline
\multirow{3}{*}{\textbf{Cognitive BCI}} 
 & Hinss2021~\cite{hinss2021eeg} & 15 & 250 & 62 & 3 & MCPS & \cite{kobler2022spd} \\
 & Covert Attention~\cite{treder2011brain} & 8 & 1000 & 62 & 6 & MCPS & \cite{fahimi2019inter} \\
 & Thinking Out Loud~\cite{nieto2022thinking} & 10 & 254 & 128 & 4 & MCPS & \cite{ng2024subject} \\
\spacedhline
\multirow{1}{*}{\textbf{Sleep Staging}} 
 & Sleep-EDF Cassette~\cite{sleepedfx_physionet} & 78 & 100 & 2 & 5 & MCPS & \cite{yang2023manydg} \\
\spacedhline
\multirow{4}{*}{\textbf{Seizure Detection}} 
 & CHB-MIT~\cite{guttag2010chbmit, shoeb2009application} & 23 & 256 & 23 & 2 & MCPS & \cite{jemal2024domain},~\cite{zhang2024cross},~\cite{zhang2025cross},~\cite{liu2025mixeeg} \\
 & PKU1st~\cite{pku1st} & 19 & 500 & 19 & 2 & MCPS & \cite{zhang2024cross} \\
 & TUSZ~\cite{Albaqami2025TUSZ} & 675 & 250 & 19* & 2 or 10 & MCPS & \cite{zhang2020adversarial},~\cite{tu2024dmnet},~\cite{wu2024invariant} \\
 & Siena Scalp~\cite{SienaScalpEEG} & 14 & 512 & 29 & 2 & MCPS & \cite{jemal2024domain} \\
 \spacedhline
\multirow{4}{*}{\textbf{Alzheimer Detection}} 
 & ADFTD~\cite{miltiadous2023dataset}& 88 & 500 & 64 & 3 & SCPS & \cite{wang2025lead} \\
 & AD-Auditory~\cite{adauditory} & 35 & 250 & 19 & 2 & SCPS & \cite{wang2025lead} \\
 & BrainLat~\cite{prado2023brainlat} & 780 & 500 & 128 & 5 & SCPS & \cite{wang2025lead} \\
& P-ADIC~\cite{shor2021eeg} & 249 & 500 & 19 & 5 & SCPS & \cite{wang2025lead} \\
\spacedhline
\multirow{2}{*}{\textbf{Multiple}} & TDBRAIN~\cite{vanDijk2022TDBRAIN, vanDijk2021TDBRAIN_Dataset} & 1274 & 500 & 26 & - & - & \cite{wang2023contrast} \\
& TUEG~\cite{obeid2016temple} & 14,987 & 250 & 64 & - & - & \cite{shen2023contrastive},~\cite{zheng2025fapex} \\
\hline
\end{tabular}}
\caption{Public EEG datasets Used in Surveyed Papers. The Label Structure indicates the mapping between subjects and classes, categorized as either Single-Class-Per-Subject (SCPS) or Multi-Class-Per-Subject (MCPS).}
\label{tab:public_eeg_datasets}
\vspace{-4mm}
\end{table*}

\subsection{Related Fields}

The solutions to the cross-subject generalization problem in EEG decoding draw heavily upon methodological frameworks developed in the broader machine learning community. The most relevant of these paradigms are Domain Adaptation and Domain Generalization, which are distinguished fundamentally by the availability of target subject data during the training phase.

\subsubsection{Domain Adaptation (DA)}
Domain Adaptation (DA) addresses the scenario where a model trained on a source domain $\mathcal{D}_S$ performs poorly on a related but distinct target domain $\mathcal{D}_T$ due to a distribution shift (i.e., $P(X_S, Y_S) \neq P(X_T, Y_T)$). The defining characteristic of DA is that it requires access to data from the specific target domain during the learning process. In cross-subject EEG decoding, this typically takes the form of \textit{Unsupervised Domain Adaptation}, where the model has access to labeled data from source subjects and unlabeled data from the target subject. The objective is to minimize the discrepancy between the source and target distributions, often by aligning their marginal distributions $P(X)$ or conditional distributions $P(Y|X)$ in a shared feature space \cite{song2023global, li2022dynamic}.

A critical distinction in EEG-based DA lies in how the source subjects are treated:
\begin{itemize}
    \item \textbf{Single-Source Domain Adaptation:} In this conventional approach, data from all available training subjects are pooled together into a single, monolithic source domain $\mathcal{D}_S = \bigcup_{i=1}^{N} S_i$. The algorithm then attempts to align this aggregated distribution with the target subject's distribution \cite{hang2019cross}. While computationally simpler, this method often ignores the inherent non-stationarity and variability among source subjects. By forcibly merging diverse neural signatures, it risks averaging out discriminative patterns, potentially creating a "confused" source distribution that aligns poorly with the target \cite{she2023multisource, yu2025fmlan}.
    
    \item \textbf{Multi-Source Domain Adaptation (MDA):} MDA explicitly recognizes that the training data originates from $N$ distinct domains (i.e., $N$ different subjects). Instead of pooling them, the model treats each source subject $S_i$ as an independent domain. Techniques in this category, such as Multi-Source Associate Domain Adaptation (MS-ADA) \cite{she2023multisource} or the Fine-grained Mutual Learning Adaptation Network (FMLAN) \cite{yu2025fmlan}, typically construct separate alignment branches for each source or learn to weigh source domains based on their similarity to the target. This granular approach allows the model to selectively leverage the most relevant subjects and avoid "negative transfer" caused by source subjects that are physiologically dissimilar to the target \cite{shi2024enhancing}.
\end{itemize}

\subsubsection{Domain Generalization (DG)}
Domain Generalization represents a more challenging and practically rigorous scenario where the target domain is completely unknown during training. Here, the model is trained on a set of source domains $\mathcal{S}_{train} = \{S_1, S_2, ..., S_K\}$ with the goal of maximizing performance on an unseen target domain $\mathcal{D}_{test}$ without accessing any of its samples. Unlike DA, which "fixes" the shift for a specific target, DG aims to learn a model that is robust to the shift itself.

Fundamentally, the core objective of Domain Generalization in broader machine learning is to learn predictive models that capture the invariant, underlying relationships between inputs and labels across multiple distinct source environments. It achieves this by extracting domain-invariant representations that represent the stable, often causal, mechanisms of a task while actively suppressing domain-specific spurious correlations that do not hold in unseen test environments~\cite{zhou2022domain, wang2022generalizing}.

Translating this paradigm to EEG decoding, the distinct "domains" naturally correspond to individual subjects~\cite{yang2023manydg}. The invariant mechanisms DG seeks to isolate are the true, task-relevant neural signatures (e.g., the neural patterns of motor imagery), whereas the spurious correlations to be suppressed are the powerful, subject-specific characteristics—such as baseline offsets or unique physiological noise profiles—that vary wildly between individuals. By leveraging DG principles to enforce stability across a heterogeneous pool of source subjects, models are prevented from overfitting to these identity-based shortcuts, theoretically ensuring robust generalization to any future, unseen subject. This capability is particularly relevant for clinical deployment, where many EEG decoding applications benefit from zero-shot generalization. Because DG methodologies do not require any target data, they enable zero-calibration systems that can function immediately for a new patient without the need for collecting any new data.

To stress the practical distinction between methodologies, we assign DA or DG to each surveyed method based on its specific data requirements prior to deployment on a new subject, as systematically displayed in Table \ref{tab:methodology_taxonomy}. DA refers to methods that require access to data—typically unlabeled—from the specific test subject during the training or adaptation/calibration phase. In contrast, DG refers to methods where the target domains (i.e., test subjects) remain completely unseen during the training process. Because DG methodologies do not require any target data, they are designed to be evaluated directly on new users without any prior calibration. 

\begin{figure*}[!htbp]
    \centering
    % First Subfigure: Task Distribution
    \begin{subfigure}[b]{0.35\textwidth}
        \centering
        \includegraphics[width=\textwidth]{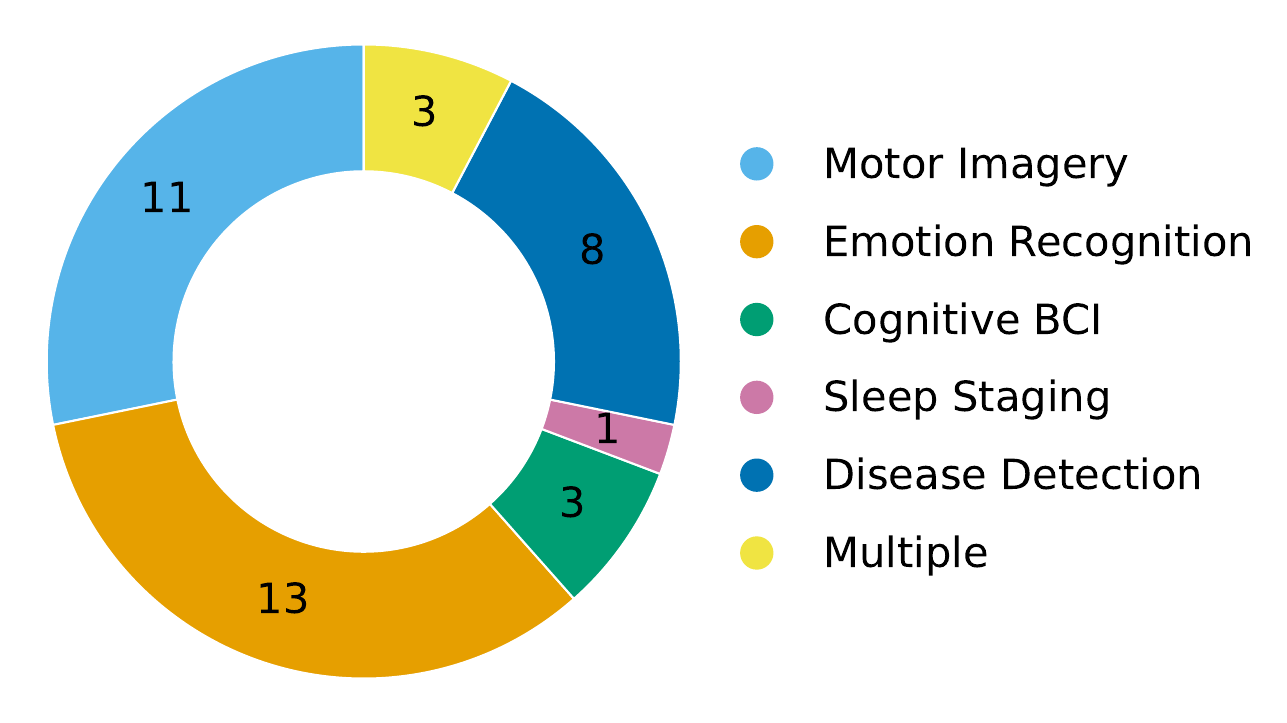}
        \caption{Task}
        \label{fig:task_dist}
    \end{subfigure}
    \hspace{-3mm}
    % Second Subfigure: Framework Distribution
    \begin{subfigure}[b]{0.27\textwidth}
        \centering
        \includegraphics[width=\textwidth]{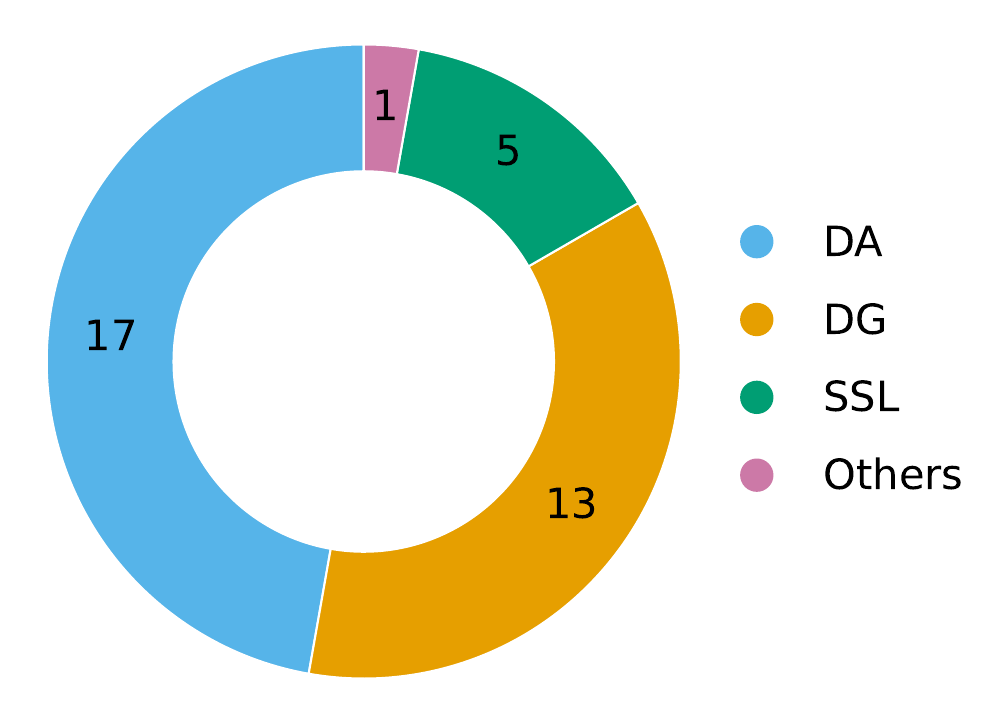}
        \caption{Framework}
        \label{fig:framework_dist}
    \end{subfigure}
    \hspace{-2mm}
    % Third Subfigure: Category Distribution
    \begin{subfigure}[b]{0.27\textwidth}
        \centering
        \includegraphics[width=\textwidth]{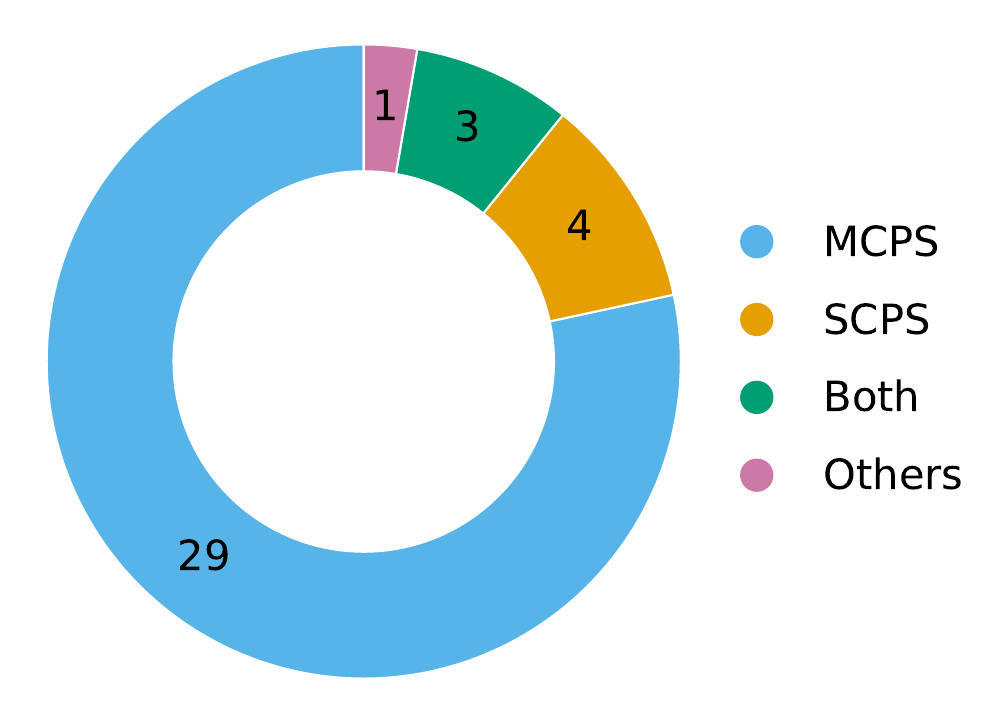}
        \caption{Label Structure}
        \label{fig:category_dist}
    \end{subfigure}
    \vspace{-2mm}
    \caption{Overview of methodology distributions.}
    \label{fig:overall_distributions}
    \vspace{-4mm}
\end{figure*}

\section{Methodological Taxonomy}
\label{sec:method}
To address the challenge of cross-subject generalization in EEG decoding, the research community has developed diverse methodologies, from feature alignment to causal representation learning. This section provides a taxonomy of these approaches, categorized by their underlying strategies for mitigating inter-subject variability. Fig.~\ref{fig:overall_distributions} illustrates the macro-level distribution of the surveyed literature, highlighting the current research emphasis across different application tasks, learning frameworks, and classification structures. Notably, the literature focuses primarily on Multi-Class-Per-Subject (MCPS) tasks such as emotion recognition and motor imagery, alongside a heavy reliance on Domain Adaptation (DA) and Domain Generalization (DG) frameworks. To unpack the algorithmic nuances driving these broader trends, a comprehensive overview of this taxonomy is presented in Table \ref{tab:methodology_taxonomy}. This table categorizes the literature into distinct methodological families—including Feature Alignment, Adversarial Learning, and Contrastive Learning—while detailing their specific mechanisms and settings (e.g., Domain Adaptation vs. Domain Generalization). The following subsections analyze each family in depth, elucidating how different methods aim to extract robust features from heterogeneous subject populations.

\begin{table*}[ht]
\centering

\resizebox{\textwidth}{!}{%
\begin{tabular}{l|l|c|l|l}
\spacedhline
\textbf{Methodological Family} & \textbf{Method / Framework} & \textbf{Setting} & \textbf{Core Mechanism} & \textbf{Specific Technique} \\ \spacedhline
\multirow{8}{*}{\textbf{Feature Alignment}} 
 & DDAN~\cite{hang2019cross} & DA & Statistical Matching & Maximum Mean Discrepancy (MMD) \\
 & Sandwich~\cite{wei2025sandwich} & DA & Statistical Matching & Federated MMD \\
 & Cross-Subject MI ~\cite{zheng2025cross} & DG & Statistical Matching & Correlation Alignment (CORAL) \\
 & SPD-BatchNorm ~\cite{kobler2022spd} & DA & Geometric Alignment & Riemannian Manifold Centering \\
 & OPS ~\cite{xu2020cross} & DA & Geometric Alignment & Online Riemannian Mean Update \\
 & MS-Manifold ~\cite{she2023cross} & DA & Geometric Alignment & Grassmann Manifold Projection \\
 & Microstate STM ~\cite{zhang2023cross} & DA & Prototype Alignment & Style Transfer Mapping (Affine) \\
 & DS3TL ~\cite{jiang2024deep} & DA & Distribution Alignment & Entropy Minimization + Pseudo-labeling \\ \spacedhline

\multirow{7}{*}{\textbf{Adversarial Learning}} 
 & DANN/CDAN ~\cite{jemal2024domain} & DA & Domain Confusion & Gradient Reversal Layer \\
 & TDANN ~\cite{bao2021two} & DA & Hybrid & MMD + Domain Discriminator \\
 & GAT ~\cite{song2023global} & DA & Hybrid & Adversarial + Center Loss \\
 & Adversarial Inf. ~\cite{ozdenizci2020learning} & DG & Subject-invariance & Gradient Reversal Layer \\
 & PANN ~\cite{zhang2024cross} & DG & Subject-invariance & Patient-Adversarial Min-Max \\
 & Confusing Loss ~\cite{hwang2020subject} & DG & Subject-invariance & Randomized Subject Labels \\
 & Deep Metric Adv ~\cite{alameer2024cross} & DG & Hybrid & Adversarial + Semantic Metric Loss \\ \spacedhline

\multirow{2}{*}{\textbf{Feature Disentanglement}} 
 & AR-Log ~\cite{zhang2020adversarial} & DG & Additive Decomposition & Two-stream Architecture \\
 & ManyDG ~\cite{yang2023manydg} & DG & Orthogonal Projection & Latent Space Orthogonality \\ \spacedhline

\multirow{5}{*}{\textbf{Contrastive Learning}} 
 & CLISA ~\cite{shen2023contrastive} & SSL & Subject-invariance & Task-level contrastive \\
 & CLOCS ~\cite{kiyasseh2021clocs} & SSL & Subject-discriminative & Subject-level contrastive \\
 & COMET ~\cite{wang2023contrast} & SSL & Subject-discriminative & Multi-level (Trial + subject) contrastive \\ 
 & LEAD ~\cite{wang2025lead} & SSL & Subject-discriminative & Sample- and subject-level contrastive \\
 & FAPEX ~\cite{zheng2025fapex} & SSL & Subject-discriminative & Sample- and subject-level contrastive \\
 \spacedhline

\multirow{4}{*}{\textbf{Ensemble \& Reweighting}} 
 & FMLAN ~\cite{yu2025fmlan} & DA & Ensemble & Teacher-Student Distillation \\
 & CANet ~\cite{zhang2025cross} & DA & Sequential & Memory Replay (Seizure Bank) \\
 % & TrAdaBoost ~\cite{feng2022classification} & SDA & Reweighting & Iterative Instance Weighting \\
 & MS-ADA ~\cite{she2023multisource} & DA & Reweighting & Source Selection via Similarity \\ \spacedhline

\multirow{3}{*}{\textbf{Subject-wise Normalization}} 
 & BCM ~\cite{kwak2023subject} & DA & Baseline Subtraction & Resting-State Feature Subtraction \\
 & InvBase ~\cite{ahmed2023novel} & DA & Spectral Filtering & Inverse Baseline Filtering \\
 & DMNet ~\cite{tu2024dmnet} & DG & Self-Comparison & Contextual/Temporal Differencing \\ \spacedhline

\multirow{2}{*}{\textbf{Transfer Learning}} 
 & CNN Fine-Tuning~\cite{fahimi2019inter} & SDA & Parameter Update & Supervised Fine-Tuning \\
 & Seegnificant~\cite{mentzelopoulos2024neural} & SDA & Parameter Update & Shared Trunk + Subject Heads \\ \spacedhline

\multirow{3}{*}{\textbf{Meta-Learning}} 
 & MTL~\cite{li2022cross} & SDA & Meta-Learning & Gradient-based Adaptation \\
 & Subj-Indep Meta~\cite{ng2024subject} & DG & Optimization & Subject-invariant Initialization \\
 & SI-Sampling~\cite{bhosale2022calibration} & DG & Sampling & Subject-Independent Sampling \\ \spacedhline

\multirow{2}{*}{\textbf{Data Augmentation}} 
 & FBGAN ~\cite{zhang2023subject} & DA & Generation & Target-conditioned GAN \\
 & mixEEG ~\cite{liu2025mixeeg} & DG/DA & Interpolation & Federated Mixup (Channel/Freq) \\ \spacedhline

\multirow{2}{*}{\textbf{Causal Learning}} 
 & Invariant Rep ~\cite{wu2024invariant} & DG & Causal Learning & Invariant Risk Minimization (IRM) \\
 & BrainOOD ~\cite{xu2025brainood} & DG & Causal Graph & Graph Information Bottleneck \\ \hline
\end{tabular}}
\caption{Categorization of Cross-Subject Methodologies in EEG Decoding. The learning settings are denoted as follows: DA (Domain Adaptation), DG (Domain Generalization), SDA (Supervised Domain Adaptation), and SSL (Self-Supervised Learning).}
\label{tab:methodology_taxonomy}

\vspace{-3mm}

\end{table*}

\subsection{Feature Alignment}
\label{sub:alignment}
The most direct approach to addressing cross-subject variability is to explicitly minimize the statistical discrepancy between source and target distributions in a shared feature space (see Fig.~\ref{fig:domain_alignment}). This methodology treats domain shift as a distributional mismatch that can be corrected by forcing the statistical moments (e.g., mean, covariance) or geometric structures of the source and target data to overlap.

\subsubsection{Statistical Moment Matching}
Common metrics for this alignment include Maximum Mean Discrepancy (MMD)~\cite{long2015learning} and Correlation Alignment (CORAL)~\cite{sun2016deep}.
MMD is a kernel-based statistical metric that measures the distance between the means of two distributions in a high-dimensional feature space to ensure their global statistics are indistinguishable. CORAL focuses on second-order statistics by aligning the covariance matrices of the source and target distributions to match the relationships between different feature dimensions. Hang et al.~\cite{hang2019cross} propose the Deep Domain Adaptation Network (DDAN), which integrates MMD into a deep convolutional network to minimize the discrepancy of deep features between subjects. Similarly, Wei et al.~\cite{wei2025sandwich} employ MMD within a federated ``Sandwich'' framework to align user-specific feature extractors with a shared central network. To capture mutually invariant representations across diverse source subjects, Zheng et al.~\cite{zheng2025cross} utilize CORAL to align the second-order statistics (covariance matrices) between every pair of source subdomains, ensuring the learned features are robust to the distribution shifts inherent between different individuals. Addressing the semi-supervised scenario, Jiang et al.~\cite{jiang2024deep} propose DS3TL, which employs an entropy-based domain adaptation module; by minimizing the prediction uncertainty (entropy) on the unlabeled target subject data, the model implicitly aligns the target distribution with the source decision boundaries.

\subsubsection{Geometric and Prototype Alignment}
Beyond Euclidean statistics, alignment can be performed on geometric manifolds or via prototype matching. Kobler et al.~\cite{kobler2022spd} introduce domain-specific batch normalization on the Riemannian manifold of Symmetric Positive Definite (SPD) matrices to remove geometric bias. Xu et al.~\cite{xu2020cross} extend this with an Online Pre-alignment Strategy (OPS), recursively updating the Riemannian mean using incoming test data. Manifold learning is also effective for granular alignment; She et al.~\cite{she2023cross} propose a framework that projects data onto a Grassmann manifold to preserve geometric structure during transfer. Furthermore, Zhang et al.~\cite{zhang2023cross} introduce a Style Transfer Mapping (STM) approach combined with microstate analysis; employing a Nearest Prototype Transfer strategy, they learn an affine mapping to align the source and target domains by minimizing the distance between their respective class prototypes.

\begin{figure}[t]
    \centering
    % First subfigure
    \begin{subfigure}[b]{0.48\textwidth}
        \centering
        \includegraphics[width=\textwidth]{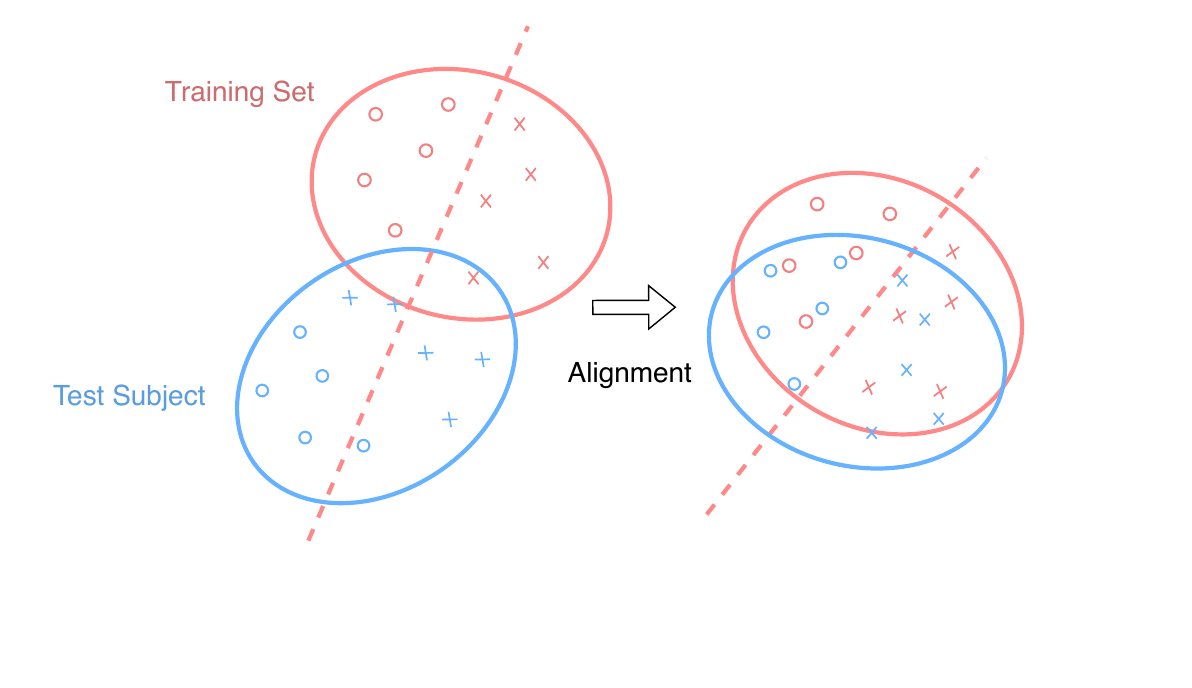}
        \vspace{-10mm}
        \caption{Domain Alignment.}
        \label{fig:domain_alignment} 
    \end{subfigure}
    \hspace{-8mm}
    % Second subfigure
    \begin{subfigure}[b]{0.48\textwidth}
        \centering
        \includegraphics[width=\textwidth]{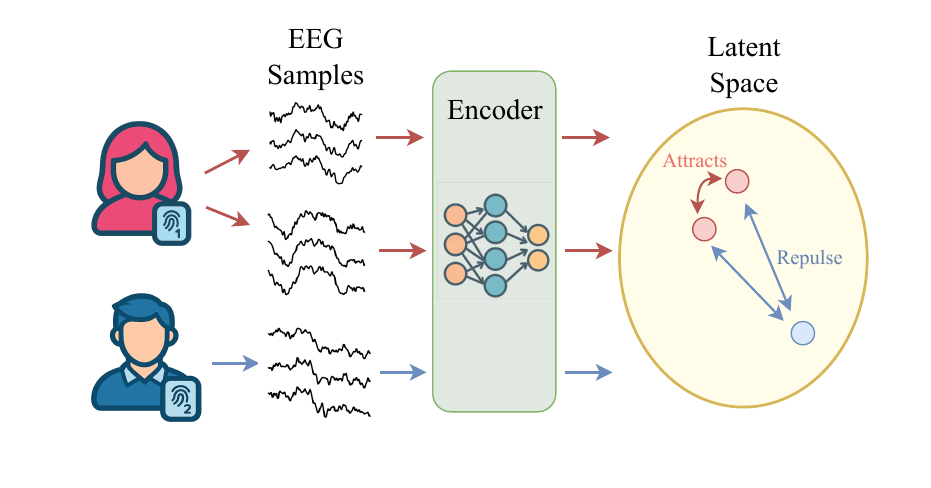}
        \caption{Subject-level Contrastive Objective.}
        \label{fig:sub_contrast}
    \end{subfigure}
    \vspace{-1mm}
    \caption{Domain alignment and subject-level contrast.}
    \vspace{-4mm}
    \label{fig:combined_methodologies}
\end{figure}

\subsection{Adversarial Learning}
\label{sub:adversarial}
Adversarial learning is a powerful technique that uses a ``minimax'' game to learn robust representations. Beyond the famous Generative Adversarial Network (GAN)~\cite{goodfellow2014gan} for generative tasks, the Domain-Adversarial Neural Network (DANN)~\cite{ganin2015unsupervised} adapts adversarial principles specifically for representation learning. The DANN framework is designed to learn features that are invariant (or agnostic) to a known nuisance variable—in this context, the subject identity.

\begin{figure}[t]
    \centering
    \includegraphics[width=0.8\linewidth]{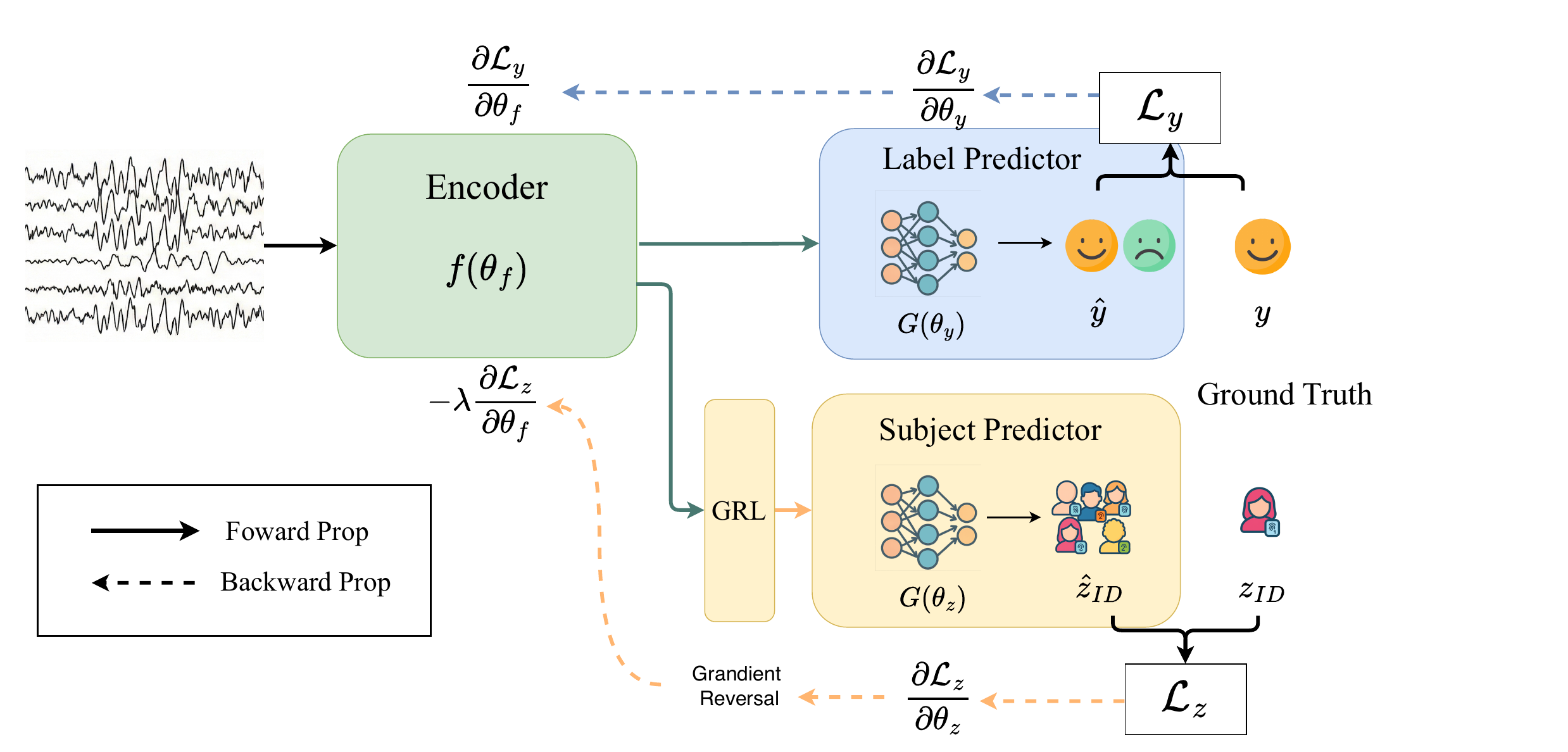}
    \vspace{-2mm}
    \caption{Domain Adversarial Neural Network.}
    \vspace{-3mm}
    \label{fig:dann}
\end{figure}

The goal is to train a feature extractor that minimizes task error while simultaneously maximizing the error of a subject discriminator. This competitive process forces the extractor to learn representations that are predictive of the task but contain no discernible subject-specific information. The most common implementation utilizes a Gradient Reversal Layer (GRL)~\cite{ganin2015unsupervised}, which inverts the gradients flowing from the subject discriminator during backpropagation (see Fig.~\ref{fig:dann}).

Several works utilize this framework to enforce systemic invariance to inter-subject variability. Özdenizci et al.~\cite{ozdenizci2020learning} employ the standard GRL approach to purge subject-specific information from EEG features. Zhang et al.~\cite{zhang2024cross} propose the Patient-Adversarial Neural Network (PANN), which adopts an alternate strategy: explicitly training the extractor to minimize the \textit{negative} of the identity classification loss. Similarly, Hwang et al.~\cite{hwang2020subject} introduce a ``confusing loss,'' training the discriminator on both real and randomized subject labels to actively prevent the model from learning subject-distinguishing features.

Recent hybrid architectures combine adversarial learning with statistical or metric constraints. Jemal et al.~\cite{jemal2024domain} systematically evaluate Conditional Domain Adversarial Networks (CDAN) for seizure prediction, demonstrating that conditioning the discriminator on the class prediction helps align complex multimodal structures. Hybrid approaches often integrate statistical alignment; for instance, Bao et al.~\cite{bao2021two} propose a two-level network (TDANN) that first uses Maximum Mean Discrepancy (MMD) for coarse alignment before employing a domain discriminator for fine-grained confusion. Similarly, Song et al.~\cite{song2023global} introduce the Global Adaptive Transformer (GAT), which couples an adversarial discriminator with an adaptive center loss to simultaneously align marginal and conditional distributions. Finally, Alameer et al.~\cite{alameer2024cross} augment the adversarial framework with Deep Metric Learning (DML), adding a ``semantic embedding loss'' that structures the embedding space by pulling same-class samples across subjects toward common proxies, ensuring that the learned invariant features remain semantically discriminative.

\subsection{Feature Disentanglement}
\label{sub:disentanglement}
Feature Disentanglement aims to achieve generalization by explicitly isolating distinct components of the representation. Unlike DANN-based methods which aim to learn subject-agnostic representation, feature disentanglement frameworks attempt to actively learn and \textit{separate} a single feature representation into distinct, isolated components. For the cross-subject problem, the goal is to learn a representation that is explicitly disentangled into a task-relevant (and subject-invariant) component and an identity (and subject-specific) component. The downstream classifier is then trained using only the task-relevant component, discarding the subject-specific information.

A direct implementation is \textit{additive signal decomposition}. Zhang et al. \cite{zhang2020adversarial} propose a framework for seizure detection that models the raw EEG signal $E$ as the composite of a seizure component $S$ and a patient component $P$, such that $E = S + P$. To achieve this, the network employs two parallel encoders: one explicitly trained to minimize seizure classification error, and the other trained to minimize patient identification error. Unlike standard adversarial methods that simply confuse a discriminator, this approach enforces a reconstruction constraint where the sum of the decomposed latent representations must recreate the original input ($E' = S + P$). This forces the model to isolate subject-specific features into the $P$ branch, leaving the $S$ branch as a pure, subject-invariant representation for diagnosis.

Alternatively, Yang et al. attempt to disentangle the features in latent space. \cite{yang2023manydg} introduce ManyDG, a framework that avoids separate encoders in favor of a unified feature vector $v$ that is mathematically decomposed during training. The method identifies a latent domain direction $z$ and projects the feature vector into two orthogonal components: $v_{\parallel z}$ (parallel to the domain factor) and $v_{\perp z}$ (orthogonal to the domain factor). The final task prediction is performed solely on $v_{\perp z}$, ensuring that the features used for decision-making are mathematically orthogonal to the patient identity. To guarantee that the factor $z$ truly captures the patient identity rather than random noise, the model utilizes a \textit{mutual reconstruction} objective: the domain factor from one sample and the label factor from a different sample (of the same subject) are combined to reconstruct the original features, thereby enforcing a strict structural disentanglement of subject identity from task content.

\subsection{Contrastive Learning}
\label{sub:contrastive}
Contrastive learning is a self-supervised paradigm that learns representations by structuring an embedding space to minimize the distance between ``positive pairs'' (semantically similar samples) while simultaneously maximizing the distance between ``negative pairs'' (dissimilar samples). In many common frameworks, such as SimCLR~\cite{chen2020simclr}, a positive pair is simply created by applying two different augmentations to the same data sample, thereby training the model to be discriminative at the sample-level.

For the cross-subject generalization problem, researchers have innovatively engineered this principle by moving beyond simple augmentations and leveraging the rich metadata unique to physiological datasets--specifically, the subject ID or the task labels. This adaptation allows for the creation of "subject-aware" or "task-aware" pairs to define the contrastive objective. These methods can typically be integrated into the Domain Generalization (DG) paradigm, as they learn a generalizable representation from a pool of source subjects without requiring access to the target subject's data. Interestingly, this has led to two distinct and seemingly contradictory strategies for achieving the same goal.

The first strategy aims to learn a representation that is explicitly subject-invariant by focusing on the shared task or stimulus. This approach is exemplified by Shen et al.~\cite{shen2023contrastive} in their CLISA (Contrastive Learning for Inter-Subject Alignment) framework. Inspired by Inter-Subject Correlation (ISC)---a neuroscience concept that different subjects perceiving the same stimulus have correlated neural activity---they define a positive pair as two EEG segments from \textit{different subjects} that were recorded during the \textit{same emotional stimulus} (e.g., the same video segment). A negative pair consists of segments from different stimuli. By maximizing the similarity of these cross-subject, same-stimulus pairs, the model is forced to learn an embedding space that clusters representations based on the shared emotional task, effectively factoring out the subject-specific neural signatures.

A second, counter-intuitive strategy aims to learn a representation that is explicitly subject-discriminative. This approach defines a positive pair as two samples from the \textit{same subject}, and a negative pair as two samples from \textit{different subjects} (see Fig.~\ref{fig:sub_contrast}). This creates an apparent paradox: to achieve cross-subject generalization, one might intuitively wish for the encoder to output a subject-invariant representation, as is the goal of the DANN methodologies. However, this subject-level contrastive objective does the opposite: it urges the model to learn subject-specific features in order to successfully distinguish between subjects in the embedding space. The resulting encoder is therefore explicitly subject-aware. This paradox suggests that a high-quality representation should isolate both subject and task features. By making subject identities explicit and well-separated, the model may allow downstream classifiers to more easily disentangle these 'nuisance' features from task-relevant patterns

This subject-discriminative strategy is the core of the CLOCS framework, proposed by Kiyasseh et al.~\cite{kiyasseh2021clocs}. Here, a positive pair consists of two different segments (e.g., from different times or different channels/leads) that belong to the \textit{same patient}. Consequently, segments from \textit{different patients} are treated as negative pairs. The model is then trained with a patient-specific loss that explicitly pulls intra-patient representations closer together and pushes inter-patient representations further apart. Wang et al.~\cite{wang2023contrast} extend this concept by integrating the patient-level objective into a hierarchical framework called COMET. The innovation of COMET is that this patient-level contrastive loss (where same-patient samples are positive pairs) is one of four distinct contrastive losses optimized simultaneously. It is combined with a trial-level loss (same-trial samples are positive) and two standard, subject-agnostic objectives: sample-level and observation-level consistency.

Recent frontier foundation models have scaled this subject-discriminative strategy by hybridizing it with standard sample-level objectives. Both LEAD~\cite{wang2025lead} and FAPEX~\cite{zheng2025fapex} adopt a dual-contrastive framework where the pre-training objective explicitly combines sample-level and subject-level contrastive losses. These approaches demonstrate that complementing instance discrimination with subject-level constraints effectively primes foundation models to extract generalized clinical biomarkers across diverse patient populations.

\subsection{Ensemble and Source-Reweighting Strategies}
\label{sub:ensemble}
Rather than forcing all subjects into a single aligned distribution, these methods explicitly acknowledge the heterogeneity of the training pool. They operate by training separate sub-models for distinct source domains (Ensembling) or by assigning importance weights to source samples based on their similarity to the target (Reweighting), thereby avoiding ``negative transfer'' from dissimilar subjects.

Yu et al.~\cite{yu2025fmlan} introduce the Fine-grained Mutual Learning Adaptation Network (FMLAN), which trains separate sub-networks for each source subject and distills their knowledge into a joint model via mutual learning. Addressing the temporal nature of BCI calibration, Zhang et al.~\cite{zhang2025cross} propose a continuous domain adaptation approach (CANet) that adapts to the target patient sequentially, using a ``seizure bank'' to replay similar historical samples and prevent catastrophic forgetting. Additionally, Feng et al.~\cite{feng2022classification} propose an instance-level transfer approach using TrAdaBoost, which iteratively re-weights source samples based on their classification accuracy on the target calibration set, effectively pruning irrelevant source data from the training distribution.

\subsection{Subject-wise Normalization}
\label{sec:selfcomparison}
A specialized family of methodologies addresses cross-subject variability by shifting the learning objective from absolute signal representation to \textit{relative} signal representation. These methods operate on the hypothesis that while the absolute characteristics of EEG signals (e.g., amplitude, power spectral density) vary wildly across subjects, the \textit{relative change} between a subject's ``resting state'' and ``task state'' remains consistent across the population . By explicitly conducting subject-wise normalization, i.e., comparing the target signal against a subject-specific reference, these approaches effectively subtract the subject-specific bias.

The most common implementation of this paradigm is \textit{baseline correction}, where a resting-state recording is used as a reference. Ahmed et al.~\cite{ahmed2023novel} introduce the InvBase method, which utilizes the power spectrum of the subject's resting-state EEG as an inverse filter; by dividing the task-state frequency spectrum by this baseline spectrum, they effectively ``de-blur'' the signal. Similarly, Kwak et al.~\cite{kwak2023subject} propose a Baseline Correction Module (BCM) within a deep neural network, where the network learns to explicitly estimate and subtract the subject-variant background feature using a paired resting-state input.

Recent work has extended this concept to feature-level self-referencing. Tu et al.~\cite{tu2024dmnet} propose DMNet, which replaces the separate baseline recording with a ``self-comparison'' mechanism using the signal's own temporal context. By computing a \textit{Difference Matrix} relative to neighboring segments, the model encodes the relative evolution of the signal rather than its absolute values. Finally, this approach can be applied at the \textit{statistical level} through domain-specific normalization. Kobler et al.~\cite{kobler2022spd} propose Symmetric Positive Definite (SPD) Domain-Specific Momentum Batch Normalization on the Riemannian manifold. Rather than subtracting a baseline signal, this method centers every subject's data at the Identity matrix on the manifold, effectively removing the ``geometric bias'' of the individual.

\subsection{Transfer Learning}
\label{sub:finetuning}
This category represents the classical supervised adaptation paradigm. A model is first pre-trained on a pool of source subjects to learn robust initial weights, then fine-tuned using a small amount of labeled calibration data from the target subject.

Fahimi et al.~\cite{fahimi2019inter} demonstrate that fine-tuning a pre-trained CNN with a small fraction of the target subject's data significantly outperforms zero-shot application, establishing the baseline efficacy of parameter transfer in EEG. Recently, Mentzelopoulos et al.~\cite{mentzelopoulos2024neural} demonstrated the scalability of this paradigm for stereotactic EEG (sEEG). They employ a Transformer-based ``shared trunk'' to extract global neural representations, followed by subject-specific regression heads that are fine-tuned to individual users. This separation of global and local parameters allows the model to handle the extreme heterogeneity of electrode placement in clinical settings, adapting to specific neural topographies without retraining the massive feature extractor.

\subsection{Meta-Learning}
\label{sub:meta}
Meta-Learning, often described as ``learning to learn,'' addresses the cross-subject challenge by restructuring the training process. Rather than minimizing the empirical risk over the training set, meta-learning methods simulate the test-time domain shift during the training phase. The core principle is episodic training: the model is trained on thousands of episodes where each task represents a different subject, and is optimized to maximize its ability to generalize to a new, held-out subject. This paradigm naturally supports both Domain Generalization, by learning a universally robust initialization, and Domain Adaptation, by learning parameters that are highly responsive to fine-tuning.

One approach uses bi-level optimization to find model parameters that are robust to inter-subject variability. Ng and Guan~\cite{ng2024subject} propose a subject-independent meta-learning framework that reformulates the training objective. Instead of treating tasks as different classification problems, they treat each subject as a distinct task while maintaining the same classification objective. Their method introduces a specialized meta-loss that minimizes the divergence between the global model parameters and the subject-specific optimal parameters. Crucially, they demonstrate that this framework is effective for both zero-calibration scenarios, where the meta-learned model is applied directly to unseen subjects, and few-shot scenarios, where the model is fine-tuned with minimal target data.

Building on this optimization-based approach, Li et al.~\cite{li2022cross} integrate meta-learning with connectivity features in a framework termed Meta-Transfer Learning (MTL). To address the potential instability of meta-learning on complex physiological data, they introduce a warmup stage where the feature extractor is pre-trained on source data to learn shallow semantic features before the meta-training phase begins. During the meta-training stage, they employ a quadratic gradient update method to optimize a Multi-Scale Residual Network (MSRN). Unlike standard generalization approaches, this method is explicitly designed for adaptation, where the meta-learner is trained to adapt to a target subject using a support set, effectively bridging the individual difference gap through fast, gradient-based fine-tuning.

Complementary to bi-level optimization, other approaches focus on how training episodes are constructed. Bhosale et al.~\cite{bhosale2022calibration} argue that the key to generalization lies in how the support and query sets are sampled during episodic training. They introduce a Subject-Independent (SI) sampling strategy, where the support and query samples in a training episode are strictly drawn from different subjects. This constraint forces the model to learn a metric space where samples cluster by semantic class (e.g., emotion state) rather than subject-specific biometric signatures. Their results demonstrate that this metric-learning approach can facilitate calibration-free decoding by relying solely on reference samples from a pool of other subjects, thereby achieving generalization without direct access to the target subject's distribution.

\subsection{Data Augmentation}
\label{sub:augmentation}
Adaptation-focused augmentation aims to bridge the specific gap between source and target distributions by generating synthetic data that mimics the target subject's characteristics, effectively ``filling in'' the void in the feature space between domains.

Zhang et al.~\cite{zhang2023subject} propose a Filter Bank GAN (FBGAN) that generates high-quality synthetic EEG samples conditioned on the target subject's distribution. By introducing these synthetic ``target-like'' samples into the training set, the classifier is forced to learn a decision boundary that naturally extends to encompass the real target subject. In the context of Federated Learning, Liu et al.~\cite{liu2025mixeeg} introduce ``mixEEG,'' a framework that employs tailored Mixup strategies, specifically Channel Mixup and Frequency Mixup, to generate synthetic EEG data. This approach provides clients with averaged, privacy-preserving unlabeled target-domain data and uses interpolation to bridge the distribution gap between decentralized source clients and the target subject.

\subsection{Causal Representation Learning}
\label{sub:invariant}
Unlike statistical alignment methods that force feature distributions to look identical, Invariant and Causal Representation Learning aims to discover underlying mechanisms that remain stable across environments. It is important to clarify that within this framework, the term ``causal'' is used in the structural sense in machine learning, referring to subject-invariant features that maintain a stable functional relationship with the label across environments, rather than implying the identification of physical causal mechanisms in the neuroscientific or interventional sense. These methods typically assume that data is generated by both invariant (causal) factors and variant (spurious) factors. Instead of aligning all features, they seek to isolate the invariant factors such that the optimal classifier built upon them remains constant across all subjects.

A primary approach in this category is Invariant Risk Minimization (IRM), which leverages environmental diversity to find stable features. Recent work~\cite{wu2024invariant} applies this principle to epilepsy diagnosis through a Spatiotemporal Invariant Risk Minimization (ST-IRM) framework. This method actively constructs diverse training environments by applying K-means clustering to partition the patient population into distinct groups. To extract stable features from these environments, the framework employs a learnable mask function that explicitly decomposes the EEG representation into invariant and variant components. The model is then optimized using a gradient variance penalty, which forces the predictor to perform consistently across all patient clusters, effectively filtering out subject-specific noise while retaining the robust spatiotemporal patterns of the seizure.

Complementary to this is Causal Subgraph Extraction, which is particularly effective for graph-structured brain networks. Xu et al. \cite{xu2025brainood} propose BrainOOD, a framework that addresses the challenge of out-of-distribution generalization by assuming that a stable causal subgraph exists within the noisy brain network. The method employs an improved Graph Information Bottleneck (GIB) objective to selectively filter out spurious connections and node features. By enforcing an alignment loss that encourages the selection of consistent functional connections across batches, the model recovers a sparse, causal subgraph that serves as a robust, subject-invariant biomarker for neurological disorder diagnosis.

\section{Discussion}
\label{sec:discussion}
\subsection{Practical Considerations for Method Selection}

In real-world deployment, the choice of a cross-subject EEG decoding model is primarily dictated by the practical data constraints of the specific application. While the algorithmic mechanisms detailed in Section 3 provide a structural overview of the methodology, a method’s suitability for a given constraint is best determined by its problem setting (e.g., DG, DA, SDA, or SSL), as categorized in Table \ref{tab:methodology_taxonomy}. The following analysis offers guidance on how these settings align with different data requirements and constraints.

\begin{table*}[ht]
\centering
\resizebox{\textwidth}{!}{%
\begin{tabular}{l|l|l|l}
\spacedhline
\textbf{Framework} & \textbf{Best Use Case} & \textbf{ Data Access Requirements} & \textbf{Primary Trade-off} \\ \spacedhline
\multirow{2}{*}{\textbf{Domain Adaptation (DA)}} & Known target subject; & Requires \textbf{fully labeled source datasets} and a & Subject-specific performance; \\ 
 & unlabeled target data available & to \textbf{unlabeled subset from the target subject}. & requires target data prior to inference. \\ \spacedhline
\multirow{2}{*}{\textbf{Supervised DA (SDA)}} & High-precision & Requires \textbf{fully labeled source datasets} and a & High accuracy but practically limited \\ 
 & clinical tuning & \textbf{labeled subset} from the target subject. & by its dependence on labeled target data. \\ \spacedhline
\multirow{2}{*}{\textbf{Domain Generalization (DG)}} & Zero-calibration; & Requires \textbf{fully labeled source datasets} & Harder to optimize; must learn \\ 
 & "Plug-and-play" deployment & and \textbf{no access} to target subject data. & truly universal invariant features. \\ \spacedhline
\multirow{2}{*}{\textbf{Self-Supervised Learning (SSL)}} & Large-scale pre-training; & Does \textbf{not require task labels}; requires \textbf{Subject-ID} & Requires large batches and significant \\ 
 & unlabeled repositories & and often \textbf{trial/temporal info}. & compute; features are task-agnostic.  \\ 
\hline

\end{tabular}}
\vspace{-2mm}
\caption{Comparison of Methods in Problem Settings}
\vspace{-2mm}
\label{tab:methods_comparison}
\end{table*}

As illustrated in Table~\ref{tab:methods_comparison}, the practical utility of each methodological setting is largely determined by the type and amount of data available before deployment. When no data from the target subject can be collected, \textbf{Domain Generalization (DG)} is the most appropriate setting, as it is explicitly designed for zero-calibration deployment on unseen subjects. When unlabeled data from the target subject is available prior to inference, \textbf{Domain Adaptation (DA)} becomes suitable because it can leverage this subject-specific information to reduce the distribution gap between source and target domains. In scenarios where a small amount of labeled target-subject data can be obtained, \textbf{Supervised Domain Adaptation (SDA)} is the most appropriate choice, particularly for applications that prioritize maximal subject-specific accuracy over deployment convenience. However, because collecting subject-specific target data is often costly and burdensome in real-world settings, methods that depend on such data may be less practical for broad deployment. From this perspective, \textbf{DG} methods remains the most viable, although also the most challenging in terms of performance, setting for real-world cross-subject EEG applications, since they must generalize to unseen subjects without access to target-subject data during training.

\textbf{Self-Supervised Learning (SSL)} is valuable under a different set of data constraints. First, it is well suited to cases where task labels in the training set are limited, since self-supervised representation learning can be driven by auxiliary information such as subject identity or trial correspondence. Second, SSL is useful when the dataset for a specific EEG task is too small to support effective supervised training. In such cases, SSL enables models to leverage general neural representations learned from larger EEG repositories, potentially collected for different tasks, and then transfer these representations to the task of interest. In this sense, SSL provides a practical route for improving cross-subject decoding when labeled task-specific data is scarce but broader unlabeled EEG resources are available. This is precisely the strategy adopted by EEG foundation models, which pretrain on large, unlabeled EEG corpora to learn transferable representations before adaptation to downstream tasks; we discuss this trend further in Section~\ref{sub:foundation}.

\subsection{Current Limitations in Cross-Subject EEG Research}
Despite the progress reviewed in this survey, several important limitations continue to constrain the development of robust cross-subject EEG decoding systems. These limitations arise at two levels: first, from methodological formulations that may not fully match the heterogeneous multi-subject nature of EEG data, and second, from inconsistencies in benchmarking practice that make it difficult to compare reported performance across studies.

\subsubsection{The Misplacement of Single-Source Domain Adaptation}
While Domain Adaptation (DA) has been a dominant paradigm for addressing cross-subject variability, a critical examination of the literature suggests that the \textit{Single-Source DA} framework, where all training subjects are pooled into a single domain $\mathcal{D}_S$ to be aligned with the target $\mathcal{D}_T$, may be theoretically misplaced for cross-subject EEG decoding. This critique rests on the statistical reality of the inter-subject variability established in Section~\ref{sub:variability}. The distribution shift is not merely a binary gap between ``Training Data" and ``Testing Data"; rather, it is a pervasive phenomenon that exists between \textit{any} two individuals. Quantitatively, the distribution gap between a target subject and a source subject is not necessarily larger than the gap between two different subjects within the training set~\cite{wang2024medtsevaluation, yu2025fmlan}.

Consequently, the Single-Source assumption, that the training set represents a coherent, unified distribution $P(X_S)$ that simply needs to be shifted to match the target $P(X_T)$, is flawed. Some existing works pool diverse subjects into a single source \cite{hang2019cross, song2023global, li2022dynamic}. Such Single-Source methods risk forcing the alignment of highly disparate subject distributions can inadvertently lead to \textbf{negative transfer}. Zhong et al.~\cite{zhong2024eeg} demonstrated that such naive single-source alignment could actually decrease performance, whereas their multi-source DG framework, designed to capture invariant relationships across a heterogeneous pool of subjects, achieved superior results.

Furthermore, This framing also exposes a paradox in the learning objective. If a feature encoder is powerful enough to handle the distribution shifts \textit{among} heterogeneous source subjects (i.e., mapping Subject A and Subject B into the same invariant feature space), it should in principle also be robust to the shift from source to target without explicit adaptation. The fact that these models still fail to generalize suggests that they are not truly resolving the internal distribution shifts of the source data. Instead, as recent evaluations on data leakage in SCPS tasks indicate \cite{brookshire2024data, wang2024medtsevaluation}, models may be overfitting to subject-specific identities rather than learning robust invariant features.

\subsubsection{Lack of Standardized Benchmarking Pipelines}
While the taxonomy presented in Section~\ref{sec:method} organizes the methodological landscape of cross-subject generalization, rigorous quantitative comparison among these approaches remains difficult. A major obstacle is the lack of standardized benchmarking pipelines across studies. Even when researchers use the same public datasets, differences in preprocessing choices—such as artifact rejection thresholds, spectral filtering ranges, and subject exclusion criteria—as well as differences in evaluation design, including subject split strategies and validation protocols, can substantially alter the difficulty of the task and the resulting performance estimates. Consequently, reported accuracies do not always reflect only the intrinsic merits of a proposed method, but are also shaped by the specific experimental pipeline under which the method is evaluated.

This problem is further compounded by the fact that many EEG datasets are small and noisy. Under such conditions, variations in preprocessing and evaluation pipelines can substantially affect reported performance, sometimes at a magnitude comparable to the gains attributed to newly proposed methods. Although comparisons against baselines within the same paper are typically conducted under a shared preprocessing and evaluation setup, this does not fully eliminate the problem: when performance is highly sensitive to these design choices, ad hoc pipeline selections can favor the proposed method, making its improvement over the baseline appear larger than it would under a different but equally plausible setup. The difficulty becomes even greater when comparing results across different papers, where preprocessing and evalution protocols often differ. As a result, both the reliability of reported gains within individual studies and the comparability of results across studies remain important concerns for the field.

Addressing this limitation will require more systematic benchmarking efforts. In particular, the development of unified benchmark datasets or benchmark suites—with standardized preprocessing pipelines, fixed subject-independent splits, and transparent reporting protocols—would provide a more reliable basis for evaluating methodological advances. A useful recent example is the 2025 EEG Foundation Challenge~\cite{aristimunha2025eeg}, which includes recordings from over 3,000 participants across six cognitive tasks. The challenge provides a common data format, downsampled releases, public starter kits, and a code-submission evaluation framework in which organizers run participant models for inference on the competition infrastructure, thereby reducing variation from ad hoc local evaluation setups and helping participants compete on a fairer common ground. Such benchmark efforts can help distinguish true algorithmic improvements from pipeline-dependent performance fluctuations and make comparisons across studies substantially more meaningful.

\subsection{Emerging Directions}
Beyond practical deployment considerations and current limitations, this survey also highlights several emerging directions that may shape the next stage of progress in cross-subject EEG research. These directions are not defined primarily by a single methodological family, but by broader conceptual shifts in how inter-subject variability is modeled and how transferable EEG representations are learned. In particular, recent work increasingly treats subject-level information not merely as a nuisance factor, but as useful structural metadata, while large-scale pretraining is opening new possibilities for learning general neural representations from diverse EEG corpora. Together, these trends point to several broader research directions for cross-subject EEG decoding, especially regarding the role of subject structure and data scale in EEG foundation models.

\subsubsection{Subject ID as Metadata}
An emerging direction in cross-subject EEG research is to treat Subject ID not merely as an indexing variable, but as structural metadata that can explicitly guide representation learning. In the context of EEG decoding, the mathematical formulation of inter-subject variability dictates that one subject is equivalent to one distinct domain (see Section~\ref{sub:variability},~\ref{sub:problem_statement}). Consequently, the Subject ID serves the exact same structural role as the domain label does in standard Domain Generalization (DG) frameworks. Both act as critical \textit{meta-information} that explicitly partitions the dataset into heterogeneous generating distributions. In general machine learning, DG without explicit domain labels is widely recognized as a more challenging setting~\cite{zhou2022domain}. We infer that the same principle holds true for cross-subject generalization: actively incorporating available Subject IDs into the training process provides a distinct advantage to a model's ability to learn robust, subject-invariant representations. 

The practical value of this perspective becomes especially apparent under the strict data governance common in clinical and physiological datasets. Ideally, researchers might wish to use detailed demographic, anatomical, or physiological profiles to help model inter-subject variability; however, such sensitive metadata is often removed to preserve privacy and anonymity. In contrast, the discrete, anonymized Subject ID is almost universally retained simply to separate data sources. Rather than treating this surviving \textit{meta-information} as a mere indexing artifact, future methods can instead repurpose it as a structural prior for cross-subject generalization.

Indeed, many of the dedicated cross-subject methodologies reviewed in this survey already harness Subject ID, predominantly by integrating it directly into the network's loss functions. For instance, both Adversarial Learning (Section~\ref{sub:adversarial}) and Feature Disentanglement (Section~\ref{sub:disentanglement}) frameworks typically formulate an auxiliary task of subject identification. They use Subject ID as the ground-truth label to either train a discriminator in a minimax game or explicitly segregate subject-specific biometric signatures into an isolated latent space. In subject-level contrastive methods (Section~\ref{sub:contrastive}), Subject ID acts as the foundational heuristic for structuring the embedding space, dictating the construction of positive (intra-subject) and negative (inter-subject) sample pairs. Similarly, Meta-Learning frameworks (Section~\ref{sub:meta}) rely on Subject ID to restructure the training process itself, defining each individual subject as a separate simulated learning task to optimize the model for rapid adaptability. Through these diverse mechanisms, Subject ID is elevated from a simple dataset index to a core algorithmic signal, and its more deliberate use may become an important direction for developing robust cross-subject EEG models.

\subsubsection{EEG Foundation Models}
\label{sub:foundation}

Recent advancements have seen the emergence of \textbf{EEG Foundation Models} that achieve state-of-the-art performance on cross-subject benchmarks. We argue that these models represent a developmental path that is largely orthogonal to the domain adaptation and generalization methodologies discussed in this survey. While the methods reviewed in Section 3 focus on \textit{algorithmic innovations}—designing specialized losses, alignment mechanisms, or architectures to handle distribution shifts with limited data—foundation models primarily drive performance through \textit{data scaling}. By pretraining on massive, diverse corpora of EEG data (spanning thousands of subjects and multiple datasets), these models learn robust, transferable representations simply by observing the full breadth of inter-subject variability \cite{wang2025lead,zheng2025fapex}.

Within this paradigm, we observe two distinct pretraining strategies. The first relies on general, task-agnostic objectives, most notably Masked Patch Reconstruction~\cite{jiang2024large, wang2024cbramod} (inspired by Masked Autoencoder~\cite{he2022masked} in Computer Vision). These models learn the intrinsic structure of neural dynamics by reconstructing missing segments of the EEG signal, implicitly capturing generalizable features without specific guidance on subject identity.

The second strategy explicitly incorporates the methodologies analyzed in this survey, most notably the Subject-Level Contrastive Learning discussed in Section~\ref{sub:contrastive}. Rather than treating \textit{subject identity} solely as a nuisance variable to be discarded, models in this category actively utilize it as an organizational constraint. By integrating subject-discriminative objectives—such as minimizing the distance between samples from the same subject—into the pretraining phase, these foundation models construct a latent space characterized by distinct, well-defined subject clusters \cite{wang2023contrast}. Recent evaluations indicate that this explicitly clustered structure empirically benefits the performance of downstream tasks. We anticipate that future advancements will stem from the convergence of these paradigms: integrating the subject-aware algorithmic constraints of Domain Generalization into large-scale Foundation Model pretraining.

\textbf{Parameter-Efficient Fine-Tuning.}
As EEG foundation models scale toward hundreds of millions of parameters, the traditional "pre-train then fully fine-tune" paradigm becomes increasingly impractical for clinical settings with limited data. This has led to the emergence of Parameter-Efficient Fine-Tuning (PEFT) as a leading strategy for adapting large-scale models to new subjects with minimal calibration. Rather than updating the entire network, PEFT maintains a frozen "shared trunk" of global neural representations and only optimizes a tiny fraction of subject-specific or task-specific parameters.

Several specialized strategies are now defining this state-of-the-art in efficient adaptation. FORMED~\cite{huang2026repurposing} exemplifies a modular repurposing strategy where the entire foundation backbone and a shared decoding attention module remain frozen; adaptation to a novel subject or task is achieved by updating only channel embeddings and label queries, allowing the model to handle diverse electrode configurations by training only ~0.1\% of the total parameters. Other PEFT techniques involve the use of Adapters and Low-Rank Adaptation (LoRA) to refine prior knowledge without altering base representations. For instance, the EEG-GraphAdapter~\cite{suzumura2024graph} integrates a Graph Neural Network (GNN) module as an adapter into a frozen temporal backbone to capture spatial relationships between sensors. REVE~\cite{ouahidi2025reve} utilizes LoRA to adapt its versatile embeddings to various electrode arrangements and subject signatures. By leveraging these efficiency-driven strategies, the field is moving toward a hybrid paradigm: utilizing large-scale pre-training to achieve a robust "universal" initialization, followed by rapid, minimal-parameter refinement for individual patients.

\section{Conclusion}
\label{sec:conclusion}
Cross-subject generalization remains a fundamental challenge in the application of deep learning to EEG decoding. In this survey, we review the landscape of cross-subject EEG decoding by discussing its underlying domain shift, task structures, and evaluation protocols. A primary contribution of this survey is the systematic taxonomy of existing deep learning methods, including feature alignment, adversarial learning, feature disentanglement, contrastive learning, and related paradigms. Alongside this taxonomy, we also emphasize a conceptual framing of cross-subject generalization as a \textit{multi-source domain shift} problem and distinguish between Single-Class-Per-Subject and Multi-Class-Per-Subject tasks. Based on this analysis, we discuss current limitations and emerging directions in the field. In particular, our survey suggests that domain generalization settings are especially relevant for real-world deployment because they do not require access to target-subject data during training or adaptation, and therefore naturally support zero-calibration use cases. At the same time, these settings remain challenging, and continued progress will likely depend on both improved methodological design and more rigorous evaluation practices. We hope that this survey provides useful insights for researchers and helps support further advances toward robust and practical EEG decoding systems.

\bibliographystyle{ieeetr}
\bibliography{ref}

\end{document}